\documentclass[letterpaper, 10 pt, conference]{ieeeconf}  

\IEEEoverridecommandlockouts                              

\usepackage{graphics} 
\usepackage{epsfig} 
\usepackage{mathptmx} 
\usepackage{times} 
\usepackage{amsmath} 
\usepackage{amssymb}  

\usepackage{color}
\usepackage{xcolor}
\usepackage{preambles}

\usepackage{amsmath}
\usepackage{graphicx}
\usepackage{url}
\usepackage{subcaption}
\graphicspath{ {images/} }
\usepackage{float}

\usepackage{amsthm}
\usepackage{mathtools}
\newcommand{\norm}[1]{\left\lVert#1\right\rVert}
\usepackage{algorithm}
\usepackage{algorithmicx}
\usepackage{algpseudocode}
\newtheorem{theorem}{Theorem}

\newtheorem{remark}{Remark}

\newtheorem{definition}{Definition}

\usepackage{ upgreek }
\usepackage{textgreek}

\newcommand{\x}{\mathbf{x}}

\newcommand{\uu}{\mathbf{u}}
\newcommand{\s}{\mathbf{s}}

\title{\LARGE \bf
On Safety Testing, Validation, and Characterization with Scenario-Sampling: A Case Study of Legged Robots}
\author{Bowen Weng$^{1}$, Guillermo A. Castillo$^{1}$, Wei Zhang$^{2}$, and Ayonga Hereid$^{3}$
\thanks{$^{1}$Electrical and Computer Engineering, Ohio State University, Columbus, OH, USA;  {\tt\footnotesize \{castillomartinez.2, weng.172\}@osu.edu.}}
\thanks{$^{2}$SUSTech Institute of Robotics, Southern University of Science and Technology (SUSTech), China; {\tt\footnotesize zhangw3@sustech.edu.cn.}}
\thanks{$^{3}$Mechanical and Aerospace Engineering, Ohio State University, Columbus, OH, USA. {\tt\footnotesize hereid.1@osu.edu.}}%
}

\begin{document}
\maketitle
\begin{abstract}
The dynamic response of the legged robot locomotion is non-Lipschitz and can be stochastic due to environmental uncertainties. To test, validate, and characterize the safety performance of legged robots, existing solutions on observed and inferred risk can be incomplete and sampling inefficient. Some formal verification methods suffer from the model precision and other surrogate assumptions. In this paper, we propose a scenario sampling based testing framework that characterizes the overall safety performance of a legged robot by specifying (i) where (in terms of a set of states) the robot is potentially safe, and (ii) how safe the robot is within the specified set. The framework can also help certify the commercial deployment of the legged robot in real-world environment along with human and compare safety performance among legged robots with different mechanical structures and dynamic properties. The proposed framework is further deployed to evaluate a group of state-of-the-art legged robot locomotion controllers from various model-based, deep neural network involved, and reinforcement learning based methods in the literature. Among a series of intended work domains of the studied legged robots (e.g. tracking speed on sloped surface, with abrupt changes on demanded velocity, and against adversarial push-over disturbances), we show that the method can adequately capture the overall safety characterization and the subtle performance insights. Many of the observed safety outcomes, to the best of our knowledge, have never been reported by the existing work in the legged robot literature.
\end{abstract}

\section{Introduction}\label{sec:intro}
Legged robots are expected to operate in real-world, shared with human, in a dense, dynamically changing, unpredictable, and somewhat adversarial environment. Existing legged robots are mostly made of metal material with up to 90-kg in weight, run as fast as 20 mph (about 8.9 m/s)~\cite{breckwoldt2019speedy}, and operate at semi-autonomous or even fully autonomous mode. Such robots could become severe safety hazards that endanger human lives if they are not designed carefully, and more importantly, not tested or validated properly. At the current stage, state-of-the-art research on legged robots still primarily focuses on locomotion controllers with falling over being the primarily concerned unsafe incident. Testing, validating, and characterizing the safety performance of such a robot is also the primary focus of this paper. Note that the discussed methodology can also generalize to other work domains of the legged robot involving other components such as the navigation and collision avoidance modules, yet details are beyond the scope of this paper.

For start, consider some robots we will analyze in Section~\ref{sec:case} for example. A robot falls 3 times out of 10 trials if demanded to transition from the steady walking velocity of 0.2 m/s to a desired velocity of 0.8 m/s. If faced against a push-over force of fixed magnitude and direction applied to the torso at 0.5 Hz, one robot maintains safety for at least 100 trials at the stepping speed of $-0.2$ m/s (walking backwards), 0 m/s (stepping in-place), and 0.4 m/s, but falls over at 0.2 m/s 2 times out of 5 trails. The same robot can survive a sagittal push-over force of 40 N coupled with a transverse push-over force of 25 N from its left side (for 100 trials), but falls over 5 times out of 10 trials when the same 25 N transverse force is applied from the right side. 

That is, the dynamic response of the legged robot locomotion is non-continuous and stochastic. Such behaviors get amplified with the evolved interest of learning-based methods in the field~\cite{castillo2019reinforcement,castillo2020hybrid,castillo2021robust,castillo2022reinforcement,Siekmann-RSS-20}. As a result, the intuitive notion of ``a safe legged robot" is ambiguous. The adopted methods to imply, prove, and characterize such a notion also exhibit significant diversities without any consent. Moreover, the safe set of states for a robot may also exhibit complex structures (e.g. non-convex and non-symmetry). This presents extra challenges to the safety property analysis.

\begin{figure}
\vspace{3mm}
\centering
\begin{subfigure}{.11 \textwidth}
  \centering
  \includegraphics[width=\linewidth]{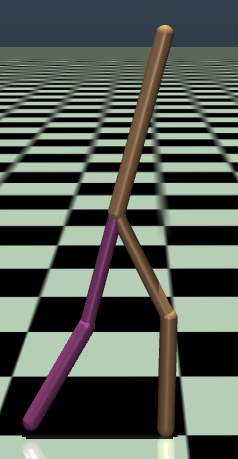}
  \caption{}
  \label{fig:rabbit}
\end{subfigure}%
\hfill
\begin{subfigure}{.14\textwidth}
  \centering
  \includegraphics[trim={0cm 0cm 0cm 1cm},clip,width=\linewidth]{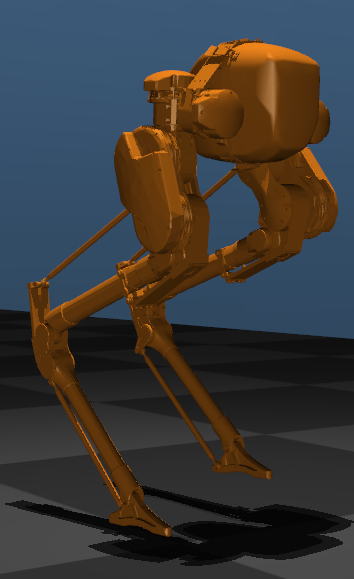}
  \caption{}
  \label{fig:cassie}
\end{subfigure}%
\hfill
\begin{subfigure}{.11\textwidth}
  \centering
  \includegraphics[width=\linewidth]{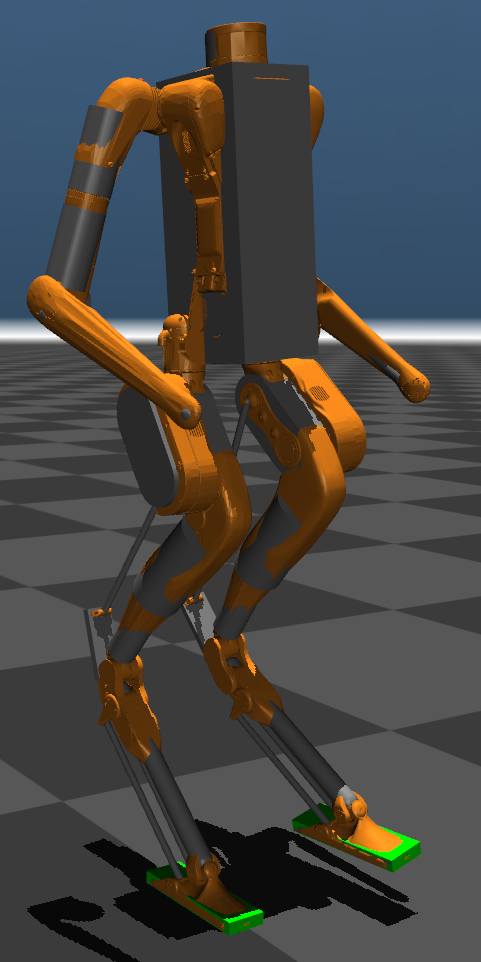}
  \caption{}
  \label{fig:digit}
\end{subfigure}
\caption{\footnotesize{The three legged robots tested for safety performance characterization in this paper: (a) Rabbit: a planar 5-DoF bipedal, (b) Cassie: a 3-D 22-DoF bipedal, and (c) Digit: a 3-D 30-DoF humanoid.}}
\label{fig:robots}
\vspace{-7mm}
\end{figure}

The most intuitive approach to justify the safety of the designed robot is to demonstrate the robot falls over with very low probability or it has been safe within a sufficiently long time period when deployed in a certain testing environment. Such an approach is essentially about observing or estimating the risk via the Monte-Carlo sampling approach. For the risk observation, low risk implies safe performance, yet such an implication is mostly intuitive without provable guarantees. For the risk inference in the automated vehicle (AV) field, Farrade et al.~\cite{fraade2018measuring} has claimed at least hundreds of millions of safe driving miles are required to demonstrate an AV's reliability in terms of fatalities and injuries. The similar scale of data should generalize to the legged robot case, which is difficult to achieve in both simulators and real-world field tests. Some have considered the importance sampling based technique~\cite{zhao2016accelerated} as a sampling efficient alternative, yet one relies on an importance function that characterizes the exposure rate of each possible event in the nominal environment. Such an important function's accuracy is difficult to justify. Moreover, simply a risk estimate is not sufficient to capture the complete safety performance of the legged robot or the subtle performance discrepancies among different robots. The risk is also propagated from an implicitly defined testing domain, hence it is not clear under what conditions the robot is safe/unsafe.

Another important direction for safety performance analysis comes from the formal verification methods~\cite{mitra2021verifying,zhao2022automated}, which rely on a certain identified system model, parametric or non-parametric, along with extended model-based analyses such as model check, reachability analysis~\cite{fan2017d}, backtracking process algorithm~\cite{hejase2020methodology}, Bayesian optimization~\cite{ghosh2018verifying}, to name a few. The method has also been extended to work in the legged robot regime~\cite{akella2022test}. The general methods of formal verification take the model of a system as its core and ``the proofs and counterexamples that come out can be only as good as the models that go in"~\cite{mitra2021verifying}. As a result, if a given model is sufficiently accurate to characterize the desired system, one can directly design provably safe robots through formal methods. Hence the verifiable safety or testing is no longer needed (as the safety property is self-guaranteed). On the other hand, if the model used to devise the robot is not sufficiently accurate, one technically cannot rely on another model for safety verification. As if the other model is sufficiently accurate, we are back to the starting point. Otherwise, one gets stuck in the infinite universe of inaccurate models. Either way, this presents us with the chicken-or-the-egg dilemma. The commonly observed model discrepancies may not necessarily be a problem for robot and algorithm designers in practice, but has restricted safety applications towards rigorous certification, unbiased assessment, benchmark comparisons, and regulatory standards.

The prevailing safety testing approach favoured by regulatory and standardization organizations takes the data-driven, scenario-based test framework with extensive applications in the software~\cite{rothermel1996analyzing}, nuclear, and automotive industry~\cite{riedmaier2020survey}. This is typically a two-step procedure, namely, the scenario-based test and the safety metric. Each scenario is a set of testing cases that characterizes the desired safety property for a certain intended functionality of the subject robot (SR). The collected data gets fed to a certain safety metric which then maps the scenario-based testing outcomes to a certain safety performance measure. Traditionally, the set of testing cases is finite and deterministic. Hence the procedure is also known as the regression test as every product iteration should regret to pass the same set of testing cases. As the test subject becomes stochastic, intelligent, and dynamically responsive, such as AV and legged robots, the regression test with surrogate and biased performance measures are no longer sufficient~\cite{hauer2020re}. The scenario-sampling based testing approach is thus considered, which also brings up new questions such as ``how many samples are sufficient?" and ``what performance measure should we use?"~\cite{weng2021towards}.

To formally characterize a robot's operational safety property, to rigorously compare the safety performance among multiple legged robots with different mechanical designs and dynamic properties, to establish a benchmark performance or a testing standard for the commercial deployment of the robot, one should not rely on exhausting field tests, a finite set of concrete scenarios, a simple scalar value of statistically inferred or observed risk, a model or a certain distribution with unjustifiable precision and impractical assumptions. This paper seeks to present a theoretically sound and sampling efficient data-driven scenario-based testing framework to characterize the safety performance of a legged robot. In particular, we generalize the notion of $\epsilon\delta$-almost safe set in~\cite{weng2021towards,weng2021formal} and adapt it to the legged robot regime as a provably unbiased safety performance measure. It rigorously specifies \emph{where} (in terms of a set of states) the robot is potentially safe and \emph{how} safe the robot is in the characterized set. Moreover, we propose a safety quantification algorithm inspired by the Synchronous Pruning and Exploration (SPE) algorithm in~\cite{weng2021towards} that efficiently converges to the $\epsilon\delta$-almost safe set of the SR. Finally, we demonstrate the performance of the proposed method with a series of state-of-the-art legged robot locomotion controllers discussed in the literature. Many of the captured safety properties, to the best of knowledge, have never been discussed in the legged robot literature before.

\textbf{Notation: } The set of real and positive real numbers are denoted by $\R$ and $\R_{>0}$, respectively. $\Z$ denotes the set of all positive integers and $\Z_N=\{1,\ldots,N\}$. The $\ell_{\infty}$-norm is denoted by $\norm{\cdot}$. $|X|$ is the cardinality of the set $X$, e.g., for a finite set $D$, $|D|$ denotes the total number of points in $D$. $|\x|$ can also denote the absolute value for some $\x \in X$. Some commonly adopted acronyms are also adopted including i.i.d. (independent and identically distributed), w.r.t. (with respect to), and w.l.o.g. (without loss of generality). 

\section{Preliminaries and Problem Formulation}\label{sec:preliminary}
In general, the manufacturer or the designer of the legged robot is expected to specify under what situations the robot should function properly without a failure for each of the robot's intended functional domain (e.g., a legged robot should not fall over on flat surface when walking at steady speed between 0 and 1 m/s). The set of such situations ($[0,1]$ m/s) is referred to as the \emph{Operational Design Domain} (ODD) of a robot. Two types of safety related problems thus arises as (i) to quantify such an ODD, or (ii) to validate if a claimed ODD is correct. This paper seeks to solve the two problems through the scenario sampling approach. For start, we establish the scenario-based testing from the dynamic system perspective as follows.

\subsection{The Scenario-based Test}
A testing system $H$ involves a SR (subject robot) along with other dynamic participants and environmental factors. It fundamentally admits a complex structure that is mostly hybrid. A testing operator controls the controllable inputs of $H$, and collects the directly observable and extracted states through a certain discrete-time data acquisition system. This leads to the motion dynamics of the following form.
\begin{equation}\label{eq:ctrl-sys}
    \s(t+1) = f_s(\s(t), \uu(t); \omega_s(t)).
\end{equation}
The state $\s \in S = X \times Q$ and $X \subset \R^n$ is a set of continuous states, $Q=\{q^1, q^2, \ldots\}$ is a set of discrete states. $Q$ can also be constructed as finite set of integers (e.g. $Q=\{1,2,\ldots\}$) for numerical calculation compatibility as we will discuss later. The action $\uu \in U$ represents the controllable input of the testing system. $\omega_s \in W_s$ and $W_s$ denotes the set of disturbances and uncertainties. Given the unknown but complex nature of $H$, $f_s: S \times U \times W_s \rightarrow S$ is in general non-Lipschitz and discontinuous. Note that the action is typically derived from a prescribed feedback control testing policy as $\uu = \pi(\s; \omega_u)$ with disturbances and uncertainties $\omega_u \in W_u$ and $\pi: S \times W_u \rightarrow U$. The testing policy can emulate the exact situation that the SR encounters in its intended functional domain (e.g. the naturalistic human driving behavior in justifying the autonomous vehicle's safety performance in the high-way driving environment). It can also represent various adversarial testing strategies~\cite{corso2019adaptive} and other desired behaviors. By replacing $\uu$ in~\eqref{eq:ctrl-sys} with $\pi(\s; \omega_u)$, we have the composed testing system dynamics as
\begin{equation}\label{eq:sys}
    \s(t+1) = f(\s(t); \omega(t)).
\end{equation}
with the composed disturbances and uncertainties $\omega \in W \in \R^w$. Let $O \subseteq S$ be a set of states that are of primary concern for a certain functionality or work domain of the robot, referred to as its operational state space (OSS). Let $\mathcal{C}$ be a set of failure states. Note that OSS and $\mathcal{C}$ are non-unique in general.

A scenario-based test thus collects and analyzes a group of sampled trajectories that characterizes the evolution of the system~\eqref{eq:sys}'s state variables within the OSS of concern. Formally speaking, a scenario is a function of the form $\sigma(K): \Z_K \rightarrow O$. 
\begin{remark}~\label{rmk:scenario}
    Note that the state evolution within a scenario starts and remains within the studied OSS unless a failure event occurs. That is, if a test scenario reaches states outside the studied OSS at a certain time $t \in \Z_{\geq1}$, we consider $\sigma(t+t')=\sigma(t-1)$ for all $t' \in \Z$ until the trajectory gets back to $O$. If a test scenario encounters a failure state at a certain time $t$ (i.e. $\sigma(t)\in\mathcal{C}$), we then have $\sigma(t+t')\in\mathcal{C}$ for all $t'\in\Z_{K-t}$. 
\end{remark}
The above configurations may not necessarily align with the ground truth state evolution. However, the detailed state trajectory is unrelated to the safety property, and the presented scenario configuration has included all necessary information for the purpose of safety study within the given OSS. A run of a scenario, $\mathcal{R}_{\sigma}(\s_0, K)$, is then defined as a sequence of acquired states $\{\sigma(i)\}_{i=1,\ldots,K}$ and $\sigma(0)=\s_0$. Note that given fixed $\s_0$ and $K$, the run of a scenario is not necessarily unique with the presence of disturbances and uncertainties in~\eqref{eq:sys}. W.l.o.g, let all scenarios be defined over the same time domain, hence $\mathcal{R}_{\sigma}(\s_0, K)$ is occasionally simplified as $\mathcal{R}_{\sigma}(\s_0)$ for the remainder of this paper.


\subsection{The Safe Set Validation and Quantification Problems}
Through consecutive runs of scenarios, this paper is primarily focused on a safety property related to characterizing and validating a certain set of states, i.e. the ODD, within which the SR is able to operate safely. The formal definition is given as follows.
\begin{definition}\label{def:odd}
The set $\Phi \subset O$ is the safe set of the SR for~\eqref{eq:sys}, also referred to as the Operational Design Domain, if $\bigcup_{\s\in\Phi} \mathcal{R}_{\sigma}(\s, K) = \Phi \cap \mathcal{C} = \emptyset$ for some given $K \in \Z$.
\end{definition}
Note that the above definition fundamentally implies $\Phi$ being a robustly forward invariant set~\cite{blanchini1999set} for~\eqref{eq:sys} with Remark~\ref{rmk:scenario}. 

The validation problem seeks to validate whether a given set is indeed the ODD by Definition~\ref{def:odd}. If the given set is fundamentally not the safe set, one only requires observing a test failure for falsification. On the other hand, if the given set is the OOD, the validation becomes a bigger challenge given the high dimension and the large coverage of OSS along with the fundamental system randomness coming from the disturbances and uncertainties. As a result, the given set is difficult to validate and the absolute safe set may not even exist. Moreover, in practice, the safe set is not always explicitly known within a given OSS. The quantification problem is thus inspired as one seeks to quantify such a potential ODD. Previous work has considered model-based solutions, such as the reachability analysis, to characterize the desired set. However, like most of the formal methods discussed in Section~\ref{sec:intro}, the model accuracy remains a challenge for unbiased justification of the safety quantification outcome. 

In the following sections, we will detail the scenario sampling based solution to tackle the above problems and adapt the solution to the application of legged robot locomotion controllers.

\section{Safe Set Validation and Quantification}\label{sec:method}

\subsection{Safe Set Validation}
As discussed above, it is impractical to claim a given set is absolutely safe. To cope with the stochastic nature of the problem, the following notion of an ``almost" safe set is adapted from~\cite{weng2021towards,weng2021formal}.
\begin{definition}\label{def:almost-safe}
    \textbf{($\epsilon\delta$-Almost Safe Set)} Given the system dynamics~\eqref{eq:sys}, $K\in\Z, \epsilon \in (0,1]$, $\delta\in\R_{\geq0}^{n+1}$, $\Phi \subseteq O$. Let $\mathcal{N}_{\delta}(\s)$ be the $\delta$-neighbourhood of $\s$, i.e., $\forall \s' \in \mathcal{N}_{\delta}(\s), |\s - \s'| \leq \delta$. The set $\Phi$ is $\epsilon\delta$-almost safe for the system~\eqref{eq:sys} if 
    \begin{itemize}
        \item For some $k \in \Z$ and $\Phi_c = \{\s_i\}_{i=1,\ldots,k}$, $\Phi_{\delta}=\bigcup_{i\in\{1,\ldots,k\}} \mathcal{N}_{\delta}(\s_i) \supseteq \Phi \supseteq \Phi_c$, and 
        \item $\mathbb{P}\Big(\big\{\forall \s \in \Phi_c: \mathcal{R}_{\sigma}(\s; K) \cap \mathcal{C}\neq \emptyset\big\}\Big)\leq \epsilon.$
    \end{itemize}
\end{definition}
That is, a set is almost safe if it is safe by Definition~\ref{def:odd} except for an arbitrarily small subset controlled by the two coefficients, $\epsilon$ and $\delta$, which are also referred to as the probability coefficient and the resolution coefficient, respectively. Note that $|\s-\s'|$ represents the absolute value and the inequality after is element-wise. Given the above definition, $\Phi_{\delta}$ is a $\delta$-covering set and $\Phi_c$ is the set of centroids. For any $\Phi$ and non-zero $\delta$, the $\delta$-covering set is non-unique. Hence a system with a given OSS may have multiple almost safe sets. As $\delta$ and $\epsilon$ tend to zero, the almost safe set becomes the safe set by Definition~\ref{def:odd}. For the set of discrete states $Q$ in particular, the corresponding value in $\delta$ as one ensures full coverage of all discrete states.

Intuitively, if one consecutively observes a sufficient number of safe runs of scenarios that start from and remain inside a given set, the set is more probably to be a safe set. Such a desired sufficiency is formally characterized by the following theorem taking advantage of Definition~\ref{def:almost-safe}.
\begin{theorem}\label{thm:validation}
    \textbf{($\epsilon\delta$-Almost Safe Set Validation)} Given the system dynamics~\eqref{eq:sys}, $\epsilon \in (0,1]$, $\beta \in (0,1]$, $\delta\in\R_{\geq0}^{n+1}$, $\Phi \subseteq O$, and the corresponding $\delta$-covering set $\Phi_{\delta}$ with centroids $\Phi_{c}$ defined by Definition~\ref{def:almost-safe}. Consider $N$ runs of scenarios, $\{\mathcal{R}_{\sigma}(\s_0^i)\}_{i=1,\ldots,N}$, with the state initialization of each run being i.i.d. w.r.t. the underlying distribution on $\Phi$ and the set of all initialization states $\{\s_0^i\}_{i=1,\ldots,N} \subseteq \Phi_c$. The set $\Phi$ is the $\epsilon\delta$-almost safe set for~\eqref{eq:sys} with confidence level at least $1-\beta$ if $\bigcup_{i=1}^N \mathcal{R}_{\sigma}(\s_0^i) \subseteq \Phi_{\delta} \cap \mathcal{C}=\emptyset$ and $N \geq \frac{\ln{\beta}}{\ln{(1-\epsilon)}}.$
\end{theorem}
One can refer to~\cite{weng2021towards} for the proof of the above theorem. This solves the validation problem and has also established the basics towards the safe set quantification solution.

\subsection{The Safe Set Quantification}
The intuitive idea of a safe set quantification algorithm is to keep sampling runs of scenarios from an instantaneous candidate set. If a scenario run reaches the failure event, a part of the set should be removed. If the exploration encounters new states inside the OSS but are outside the current candidate set. The candidate set should also be modified. Eventually, if a sufficient number of consecutive runs of scenarios start from and remain inside a candidate set, then the algorithm terminates by Theorem~\ref{thm:validation}. That is, \emph{every quantification algorithm ends up with a validation algorithm}. The computational performance of a quantification algorithm is then mainly determined by how fast it converges to a candidate set that triggers the validation process. 

In this paper, we consider Algorithm~\ref{alg:qnt} as the quantification solution. Note that \texttt{pop}, \texttt{reachable}, \texttt{norm-nearest}, \texttt{remove}, and \texttt{append} are all notional functions. $\mathcal{X}.$\texttt{pop}() returns a point $\x\in\mathcal{X}$ and removes it from the set. \texttt{reachable} ($\s, G$) returns all vertices on the graph $G$ that connects, directly and indirectly, to the point $\s$ through a depth-first-search routine. The commands \texttt{remove} and \texttt{append} simply remove a point from or add a point to the given set, respectively. $\mathcal{X}$.\texttt{norm-nearest}($\x$) returns the nearest point to $\x$ in $\mathcal{X}$ in terms of the normalized $\ell_2$-norm distance. That is, $\x$ and all points in $\mathcal{X}$ are first normalized w.r.t. the admissible value range of each individual dimension and one then propagates the $\ell_2$-norm distance between the normalized $\x$ and $\mathcal{X}$. 
\begin{algorithm}[H]
\small
    \begin{algorithmic}[1]
    \State {\bf Input:} Initial $\delta$-covering set $\Phi_{\delta} \subseteq O$ with centroids $\Phi_c$, failure set $\mathcal{C}$, $\epsilon\in(0,1]$, $\beta\in(0,1]$, $K\in\Z$.
    \State {\bf Initialize: } Graph $G_{s} = (\Phi_c, E_s), E_s =\emptyset
    \subset O^2$ and $G_u=(D_u, E_u), D_u=\emptyset\subset S, E_u = \emptyset \subset S^2$, prioritized replay buffer $\mathcal{B}=\emptyset$, N=0.
    \State{{\bf While} $N<\frac{\ln{\beta}}{\ln{(1-\epsilon)}}$:}
    \State{\ \ \ \ {\bf If} $\mathcal{B}=\emptyset$}
    \State{\ \ \ \ \ \ \ \ $\s_0 \sim U(\Phi_c)$}
    \State{\ \ \ \ {\bf Else}}
    \State{\ \ \ \ \ \ \ \ $\s_b = \mathcal{B}$.\texttt{pop}(), $\s_0 = \Phi_c.$\texttt{norm-nearest}($\s_b$)}
    \State{\ \ \ \ {\bf End If}}
    \State{\ \ \ \ Get $\tau=\mathcal{R}(\s_0, K)$}
    \State{\ \ \ \ {\bf If} $\tau\cap\mathcal{C} \neq \emptyset$}
    \State{\ \ \ \ \ \ \ \ {\bf For $i$ in $\Z_{|\tau|-1}$} {\bf do}}
    \State{\ \ \ \ \ \ \ \ \ \ \ \ $\mathcal{B}$.\texttt{append}($\tau[i]$)}
    \State{\ \ \ \ \ \ \ \ \ \ \ \ {\bf For} $\s$ in \texttt{Reachable}($\tau[i],G_s)$ {\bf do}}
    \State{\ \ \ \ \ \ \ \ \ \ \ \ \ \ \ \ $\Phi_c$.\texttt{remove}($\s$)}
    \State{\ \ \ \ \ \ \ \ \ \ \ \ $E_u$.\texttt{append}(($\tau[i]$, $\tau[i+1]$))}
    \State{\ \ \ \ \ \ \ \ \ \ \ \ {\bf End For}}
    \State{\ \ \ \ \ \ \ \ $\mathcal{B}$.\texttt{append}($\tau[i+1]$)}
    \State{\ \ \ \ \ \ \ \ {\bf End For}}
    \State{\ \ \ \ \ \ \ \ $N=0$}
    \State{\ \ \ \ {\bf Else}}
    \State{\ \ \ \ \ \ \ \ $\bar{\s}=\s_0, N_s=|\Phi_c|$}
    \State{\ \ \ \ \ \ \ \ {\bf For $i$ in $\{2,\ldots,|\tau|\}$} {\bf do}}
    \State{\ \ \ \ \ \ \ \ \ \ \ \ {\bf If} $\tau[i] \notin \Phi_{\delta}$}
    \State{\ \ \ \ \ \ \ \ \ \ \ \ \ \ \ \ $E_s$.\texttt{append}(($\bar{\s}$, $\tau[i]$))}
    \State{\ \ \ \ \ \ \ \ \ \ \ \ \ \ \ \ $\bar{\s}=\tau[i]$}
    \State{\ \ \ \ \ \ \ \ \ \ \ \ {\bf End If}}
    \State{\ \ \ \ \ \ \ \ {\bf End For}}
    \State{\ \ \ \ \ \ \ \ {\bf If} $N_s=|\Phi_c|$ and $\mathcal{B}=\emptyset$}
    \State{\ \ \ \ \ \ \ \ \ \ \ \ $N += 1$}
    \State{\ \ \ \ \ \ \ \ {\bf Else}}
    \State{\ \ \ \ \ \ \ \ \ \ \ \ $N = 0$}
    \State{\ \ \ \ \ \ \ \ {\bf End If}}
    \State{\ \ \ \ {\bf End If}}
    \State {{\bf Output:} $\Phi_{\delta}$}
    \end{algorithmic}
    \caption{Almost Safe Set Quantification} \label{alg:qnt}
\end{algorithm}

The overall flow of Algorithm~\ref{alg:qnt} has four steps as follows. (i) The initialization step (line 2) configures two graphs, $G_s$ and $G_u$, with vertices ($\Phi_c$ and $D_u$) and edges ($E_s$ and $E_u$) to hold observed states and transitions from runs of scenarios that are safe and unsafe, respectively. One also initializes an empty set $\mathcal{B}$ for the prioritized sampling as we will see later. (ii) The sampling step (line 4-9) gets an i.i.d. sample from the current $\Phi_c$ if the replay buffer $\mathcal{B}$ is empty. Otherwise (a failed run of a scenario has been observed beforehand), one gets a closed point in $\Phi_{c}$ to a popped out point from $\mathcal{B}$. The run of a scenario from the given sample is then executed. (iii) If that run ends up with a failure event (line 10-19), one moves the observed failure states and all other states connected to the observed failure ones, directly or indirectly, from $G_s$ to $G_u$. The unsafe states are also added to $\mathcal{B}$ to help future sampling near the potentially unsafe regions (see the previous step). (iv) Otherwise, the scenario run is observed safe (line 21-32). One should extend the covering set if an observed safe point is not in $\Phi_{\delta}$. If for consecutively $N$ runs of scenarios by Theorem~\ref{thm:validation}, $\mathcal{B}$ remains empty, all sampled runs of scenarios are safe and remain covered by $\Phi_{\delta}$, the algorithm terminates with the desired $\epsilon\delta$-almost safe set.

The uniform sampling in line 5 can be replaced with any distribution per the testing configuration. The prioritized sampling is an important heuristic add-on that accelerates the convergence to the almost safe set as potentially unsafe states get tested and removed more frequently. The theoretical property by Theorem~\ref{thm:validation} is not jeopardized as the replay buffer $\mathcal{B}$ should essentially be empty. The \texttt{norm-nearest} configuration also resolves the problem of a particular dimension dominating the $\ell_2$-norm distance especially for high-dimensional data with unbalanced ranges among different dimensions. 

\section{Safe Set Quantification of the Legged Robot Locomotion Controllers}\label{sec:case}
This section adapts the presented framework to work with legged robots. In the previous work by Castillo et al.~\cite{castillo2019reinforcement,castillo2020hybrid,castillo2021robust,castillo2022reinforcement}, the authors have presented a series of legged robot locomotion methods. That includes a hybrid zero dynamics (HZD)~\cite{da2019combining,gong2019feedback} inspired reinforcement learning (RL) based controller, referred to as HZDRL, for Rabbit (a 5-DoF planar bipedal robot), Cassie (a 22-DoF 3D bipedal robot), and Digit (a 30-DoF 3D humanoid) shown in Fig.~\ref{fig:robots}. For the two 3D robots, Cassie and Digit, they also have a HZD based controller with feedback regulators for performance comparison, referred to as HZD in the remainder of this paper. Moreover, Siekmann et al.~\cite{Siekmann-RSS-20} have also presented a locomotion controller for Cassie based on recurrent neural network trained with Proximal Policy Optimization (PPO), referred to as Cassie-RPPO. Although the robots are not commercially available products, they all have demonstrated competitive performance in handling complex tasks, such as tracking various speeds on different terrain conditions and maintaining stable walking gaits against push-over disturbances in both simulated and real-world tests with the same policy. However, the examples in the publications and the videos are not rigorous evidence of the robots' safety performance. Our proposed framework is thus adopted to evaluate the selected state-of-the-art legged robot locomotion controllers. 

To the best of our knowledge, there does not exist any commonly agreed OSS design for legged robot locomotion. This paper then presents four cases for performance demonstration. Note that Cassie-RPPO is a retrained model based on the open-source code and the presented methods in the original publication~\cite{Siekmann-RSS-20}. Throughout this section, all tests are conducted in the MuJoCo simulator~\cite{todorov2012mujoco} which is the primary environment where the controllers are originally developed. Note that Digit-HZDRL and Digit-HZD have also been tested in real-world (e.g. in~\cite{castillo2021robust,castillo2022reinforcement}). Theoretically and empirically, as we will see in the following case studies, the proposed scenario-sampling approach is feasible to execute in real-world with appropriate testing equipment, yet details are beyond the scope of this paper. Moreover, due to the mechanical nature of the legged robot, it is in general difficult for a robot to track a certain walking velocity accurately. Throughout this paper, the robot velocity typically represents the filtered average step velocity, which is a common setup in the field.

\subsection{Case I: tracking speed transition}
\begin{figure*}[!ht]
    \centering
    \includegraphics[trim={2cm 2cm 2cm 0cm},clip,width=0.99\textwidth]{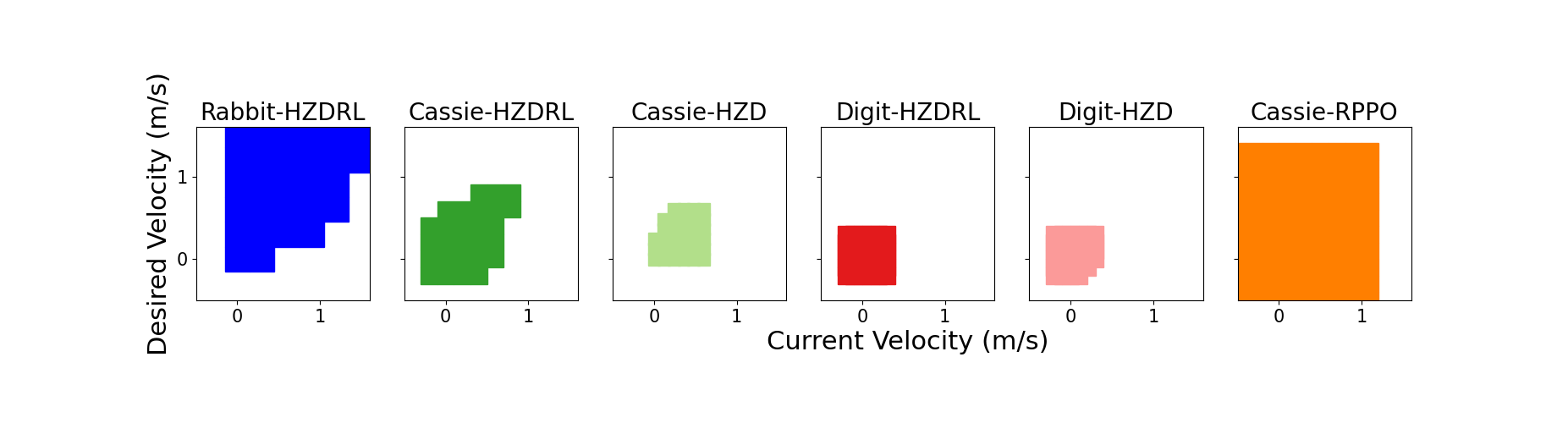}
    \caption{\footnotesize{The $\epsilon\delta$-almost safe sets for the six selected subject robots in $O_1$ ($\delta=[0.2,0.2], \epsilon=0.02, \beta=0.001$).}}
    \label{fig:sdq_speed_track}
    \vspace{-3mm}
\end{figure*}
The capability to transition form the steady-state velocity tracking status to another commanded walking speed without falling over on a flat surface is one of the basic safety properties of the legged robot. The OSS for this case admits a two-dimensional configuration as $O_1 = V_c \times V_d$ with the current tracking velocity set $V_c \subset \R$ and the desired velocity set $V_d \subset \R$, both are defined over the sagittal domain. A robot is considered at the velocity tracking status of $v_c \in V_c$ if its average step velocity at time $t$, $v(t)$, satisfies $|v_c-v(t)|\leq \bar{v}$ for a given tolerance $\bar{v} \in \R_{\geq0}$. $\bar{v}$ is set as $0.1$ for all experiments in this section. Note that the safety quantification for this OSS is only concerned with falling-over as the unsafe incident. Utility properties such as whether the robot reaches the commanded speed is not of interest. A robot will reject any desired velocity change if it is not at the velocity tracking status. We will relax this condition later in a more complex OSS example.

With $\delta=[0.2,0.2], \epsilon=0.05, \beta=0.001$, Algorithm~\ref{alg:qnt} is executed using five different random seeds for each SR with all SRs sharing the same set of seeds. For each set of five tests against the same SR, let the Intersection-over-Union (IoU) ratio be calculated based on the cardinality of the five obtained sets (IoU being one indicates that the almost safe sets are identical). Among the 6 investigated SRs, the IoU ratio remains as one for the given hyper-parameters, and decreases to a minimum ratio of 0.9 for a lower confidence level ($\beta=0.01$). Note that the non-one IoU ratio does not indicate an incorrect set as the almost safe set is fundamentally non-unique given the stochastic nature of the system. 

Examples of the obtained $\epsilon\delta$-almost safe sets for all 6 SRs sharing the same seed are shown in Fig.~\ref{fig:sdq_speed_track}. Overall, Rabbit-HZDRL can maintain safe status with high probability if demanded to accelerate to up to 1.5 m/s walking speed, but would fall over easily if asked to stop abruptly at high speeds (e.g. $v_c=1.5$ m/s and $v_d=$0.2 m/s). Cassie-HZDRL exhibits a close-to-symmetry pattern around the steady-state velocity tracking mode with $v_c=v_d$. That is, it is not safe for the robot to stop or to accelerate abruptly. Moreover, one shall note that the general notion of ``one robot is safer than the other" can be problematic. For example, Cassie-HZD and Digit-HZD have the almost safe sets with a similar size (the Cassie-HZD one is slightly bigger), yet their capabilities to handle abrupt acceleration and deceleration are completely opposite. As a result, the safety performance comparison is a multi-dimensional analysis, and the proposed method has demonstrated to be able to supply with subtle insights that facilitate the multi-dimensional analysis. Finally, Digit-HZDRL and Cassie-RPPO are the only two controllers that can accelerate and decelerate safely within their respective ODD. We then devise a modified OSS for further comparisons.

\subsection{Case II: speed transition with periodically commanded desired velocity}
The second OSS extends the design of $O_1$ by considering not only the speed tracking stage, but also the transitioning stages involving acceleration and deceleration. The robot is placed in a testing environment with the desired speed command following a periodic squared wave switching between the upper and lower bounds of each robot's trackable velocity (obtained from Fig.~\ref{fig:sdq_speed_track}) at various configurable frequencies. This leads to a three-dimensional OSS $O_2 = V_s \times F_c \times Q_c$ with the set of current step velocity $V_s \in \R$, the set of frequencies $F_c \subset \R_{>0}$, and the set of modes $Q_c = \{\texttt{deceleration}, \texttt{steady-state}, \texttt{acceleration}\}$. $\texttt{steady-state}$ denotes the velocity tracking status mentioned above, $\texttt{deceleration}$ and $\texttt{acceleration}$ are deemed by the signs of the average step acceleration. The run of such a scenario requires to initialize the test at an arbitrarily given velocity at a certain mode in $Q_c$. This can be achieved through added heuristics (having the robot tracking a series of speeds with the desired initialization in between) with very few exceptions such as the combination of the maximum velocity with a non-steady-state mode, which should not be part of the OSS anyway.

Note that the above OSS design is very similar to the standard testing of automobile's Electronic Stability Control (ECS) system~\cite{national2007fmvss} released by the the U.S. Federal Motor Vehicle Safety Standards (FMVSS). The test controls the vehicle's steering angle following a prescribed sinusoidal pattern at various steady-state longitudinal speeds to evaluate the roll-over risk of the vehicle. Intuitively, the vehicle should be unstable within a certain frequency interval as an extremely low frequency leads to sinusoidal curve following and a significantly high frequency leads to the straight driving trajectory. A similar observation generalizes to the OSS design of $O_2$ for the legged robot as well. 

\begin{figure}[t]
    \centering
    \includegraphics[trim={1cm 0cm 2cm 0cm},clip,width=0.48\textwidth]{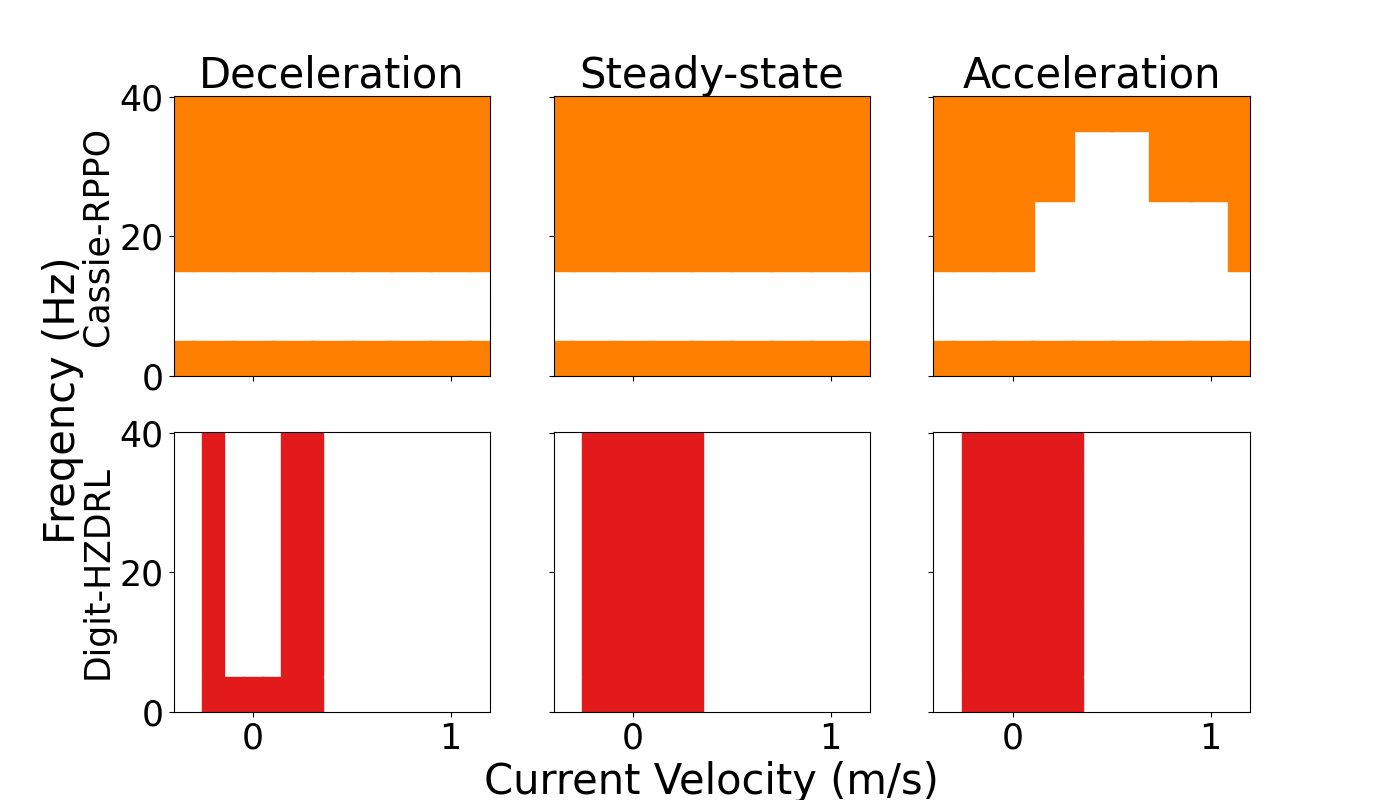}
    \caption{\footnotesize{The $\epsilon\delta$-almost safe sets for Digit-HZDRL and Cassie-RPPO in $O_2$ ($\delta=[0.2,0.5,1], \epsilon=0.02, \beta=0.001$).}}
    \label{fig:freq}
    \vspace{-5mm}
\end{figure}
\begin{figure}[t]
    \centering
    \includegraphics[trim={1cm 0cm 2cm 1cm},clip,width=0.45\textwidth]{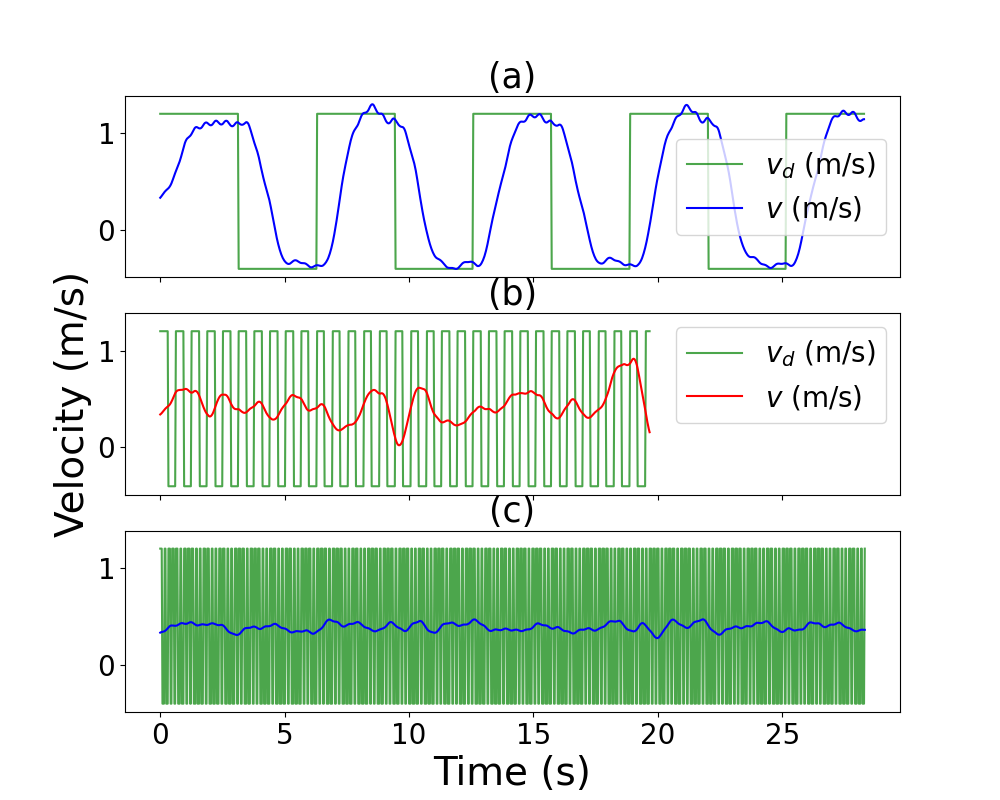}
    \caption{\footnotesize{Examples of runs of scenarios at different desired velocity changing frequencies in $F_c$ with Cassie-RPPO operating in $O_2$, the robot falls over in (b), but remains safe in (a) and (c).}}
    \label{fig:freq_demo}
    \vspace{-5mm}
\end{figure}

As illustrated in Fig.~\ref{fig:freq}, Digit-HZDRL and Cassie-RPPO are tested in the given OSS. With $\delta=[0.2,0.5,1], \epsilon=0.05, \beta=0.001$, and the same set of 5 random seeds. The IoU ratio for each SR remains as one. For Cassie-RPPO, the robot exhibits an intermediate unsafe frequency region for all modes at all velocities. The cause of such a phenomenon is empirically investigated with added runs of scenarios shown in Fig.~\ref{fig:freq_demo}, where the robot tracks the desired velocity change safely at a low-frequency (Fig.~\ref{fig:freq_demo}(a)), remains at the current velocity at a high-frequency (Fig.~\ref{fig:freq_demo}(c)), but falls over in between (Fig.~\ref{fig:freq_demo}(b)). There are also other safety properties revealed through Fig.~\ref{fig:freq_demo} such as the triangular unsafe region for Cassie-RPPO in the \texttt{acceleration} mode, and the observation that Digit-HZDRL is only unsafe within a subset of the \texttt{deceleration} mode. However, the primary purpose of the proposed safety evaluation framework is to characterize the safety property of the legged robot for regulatory, comparison, and benchmark purposes. One shall rely on fault analysis and diagnostic techniques~\cite{hejase2020methodology} for further investigation towards detailed causes of the discovered safety properties. Finally, to the author's knowledge, Digit's on-board locomotion controller provided by Agility Robotics Inc. also suffered from the revealed problem (i.e. frequently moving the speed control joystick back-and-forth may cause the robot to fall over) with the physical robot in real-world environment. The problem has been seemingly fixed with the recent firmware updates.

\subsection{Case III: tracking speed transition on the sloped surface}
\begin{figure}
    \centering
    \includegraphics[trim={0cm 0cm 2cm 0cm},clip,width=0.49\textwidth]{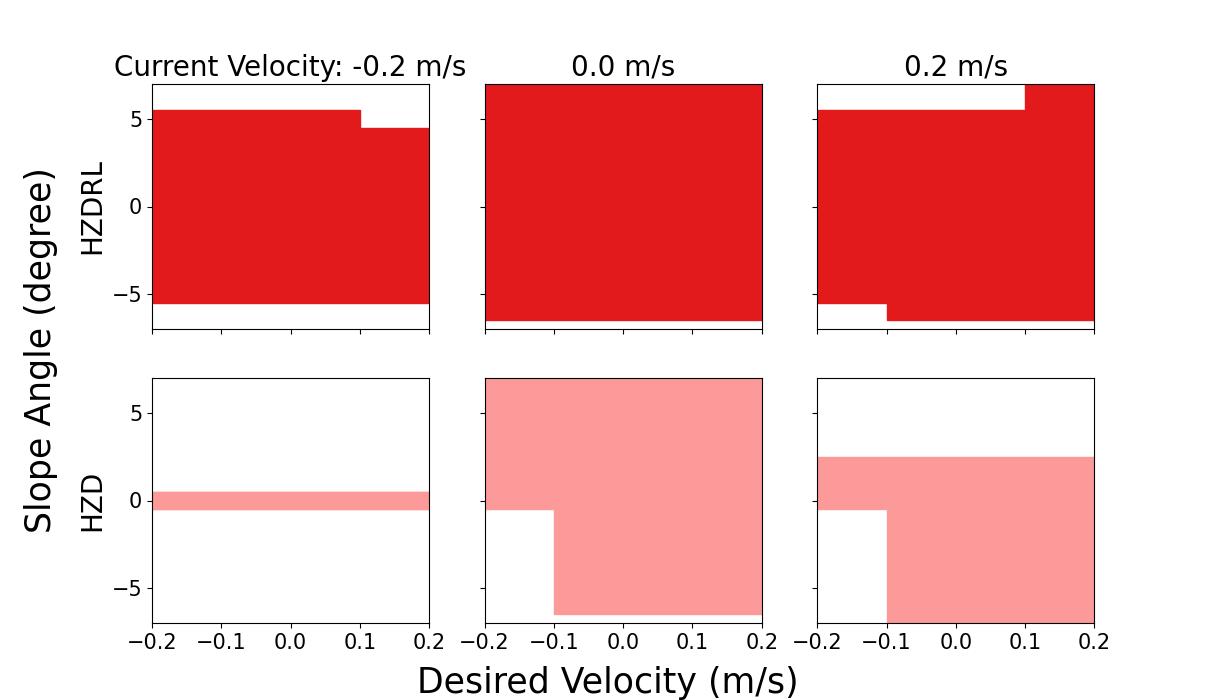}
    \caption{\footnotesize{The $\epsilon\delta$-almost safe sets for Digit-HZDRL and Digit-HZD in $O_3$ ($\delta=[0.2,0.2,1], \epsilon=0.02, \beta=0.001$).}}
    \label{fig:digit_slope}
    \vspace{-3mm}
\end{figure}

The third OSS studies the safety performance of the legged robot controller on the sloped surface. The OSS admits a three-dimensional design as $O_3=V_c \times V_d \times A$ with the set of current velocity in steady-state tracking mode $V_c$ , the set of desired velocities $V_d$ (both $V_c$ and $V_d$ are similar to the ones used in $O_1$) and the surface slope angle $A \subset \R$. The obtained $\epsilon\delta$-almost safe sets for the two Digit-based SRs are shown in Fig.~\ref{fig:digit_slope}. In this particular comparison, Digit-HZDRL is safer than Digit-HZD as the obtained almost safe set for Digit-HZD is only a proper subset of the one obtained for Digit-HZDRL. 

\subsection{Case IV: steady-state speed tracking with periodic adversarial push-over forces}
The last case is concerned with adversarial push-over forces applied to the center of gravity of the robot torso at the steady-state speed tracking mode. In particular, we consider the OSS as $O_4=V_c \times F_x \times F_y$ with the current velocity set $V_c$, the set of sagittal push-over forces $F_x \in \R$, and the set of transverse push-over forces $F_y \in \R$. The combined force is applied at 0.5 Hz.

\begin{figure}[!t]
    \centering
    \includegraphics[trim={0cm 0cm 1cm 1cm},clip,width=0.45\textwidth]{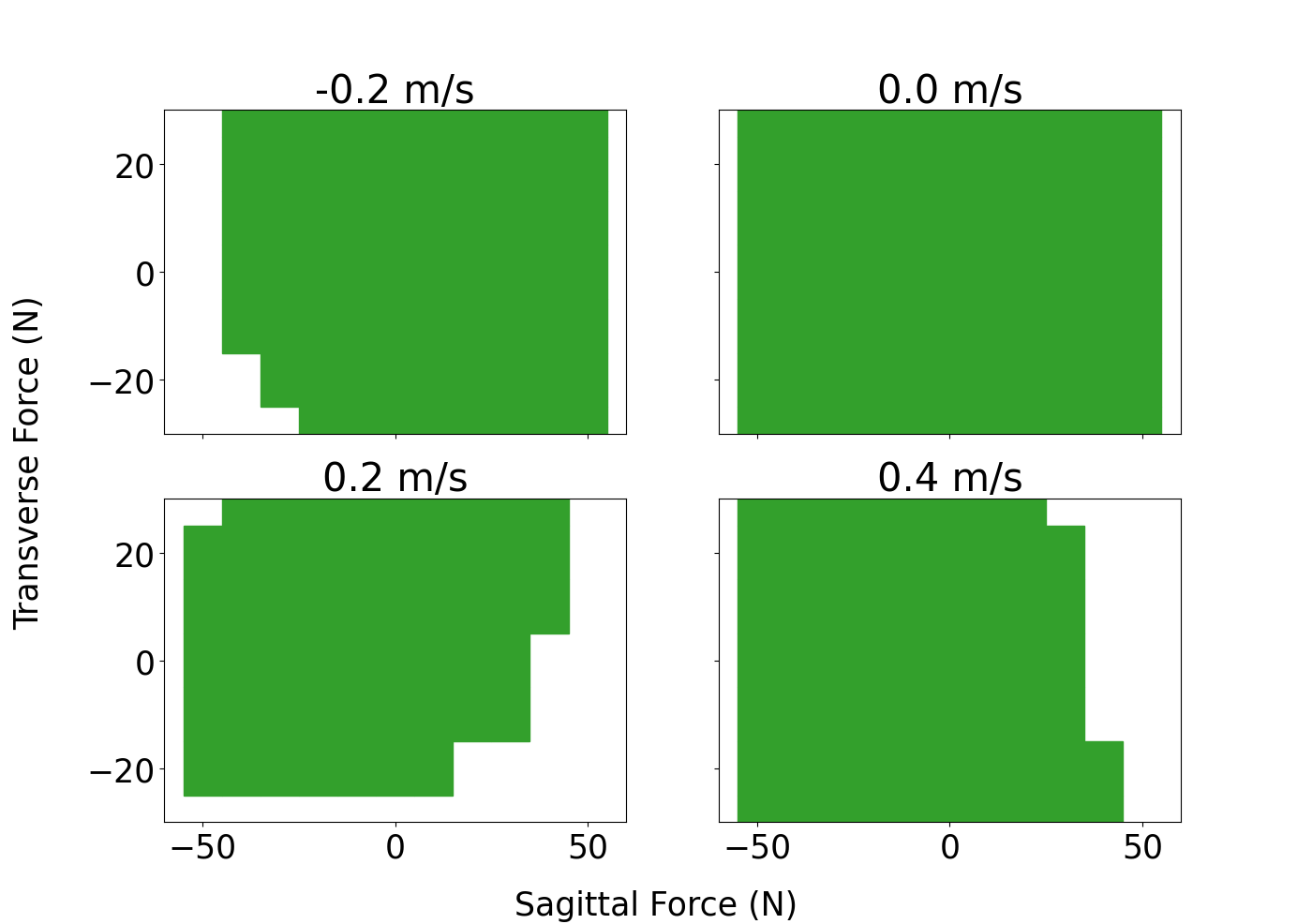}
    \caption{\footnotesize{Some subspace slicing of the $\epsilon\delta$-almost safe sets for Cassie-HZDRL in $O_4$ ($\delta=[0.2,5,5], \epsilon=0.02, \beta=0.001$).}}
    \label{fig:cassie_adv}
    \vspace{-4mm}
\end{figure}

\begin{figure}
    \centering
    \includegraphics[trim={0cm 0cm 2cm 1cm},clip,width=0.45\textwidth]{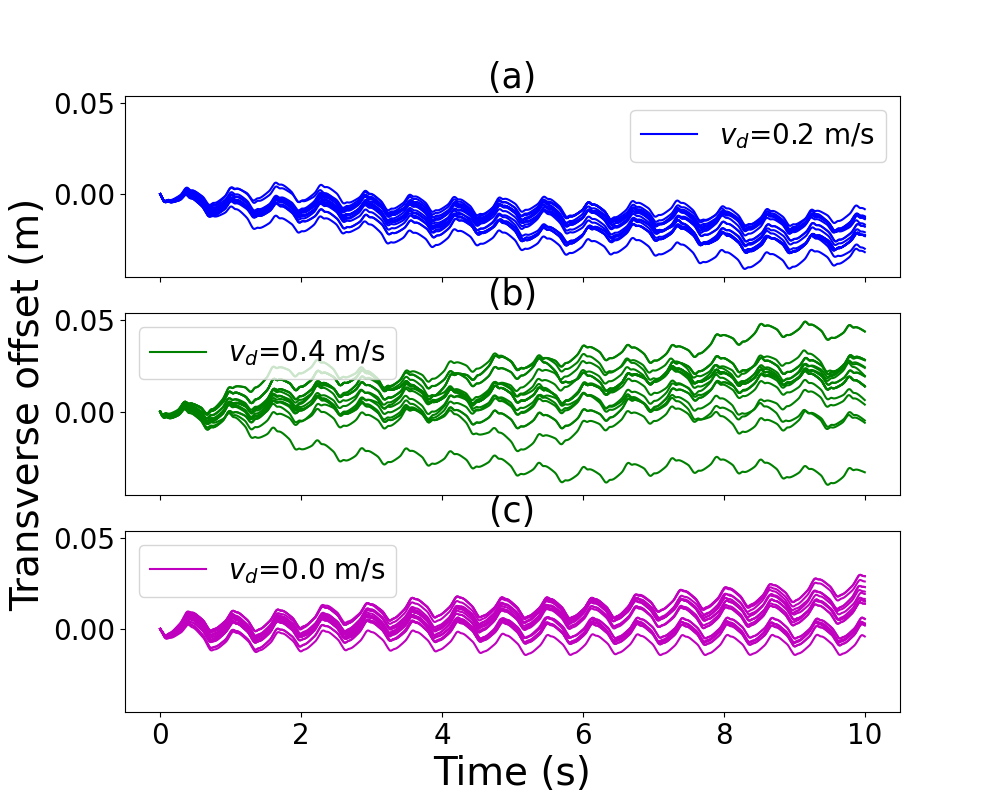}
    \caption{\footnotesize{The transverse offset distance of Cassie-HZDRL walking at various velocities in steady-state mode on flat surface without the presence of external push-over forces over 20 runs of scenarios (time 0 denotes the first time the robot reaches the desired speed).}}
    \label{fig:drift}
    \vspace{-6mm}
\end{figure}

Ideally, one might expect the safe set to be symmetric about the straight walking direction on the sagittal domain. This is not the case as shown in in Fig.~\ref{fig:cassie_adv} for Cassie-HZDRL. The SR is more stable against the left-side disturbances when walking at 0.2 m/s and against the right-side disturbances when speeding up to 0.4 m/s. A detailed investigation regarding the potential cause of such an anti-intuitive property is shown in Fig.~\ref{fig:drift}. Empirically within the transverse domain, the SR is mostly biased towards its left side when walking at 0.2 m/s and is primarily drifting towards the right side when walking at 0.4 m/s. At 0 m/s (walking in-place), the robot keeps its transverse drift small without a particular tendency towards either side. Note that the overall drift distance is very minor (less than 5-cm in 10 seconds) for all illustrated experiments, yet the subtle bias clearly affects the robot's safety performance. This presents some empirical insights that explain the almost unsafe set asymmetry observed in Fig.~\ref{fig:cassie_adv}.


\subsection{Discussions}
We conclude this section with some added remarks on sampling efficiency, quantification initialization, and the interchangeable role between OSS and the underlying disturbances and uncertainties.

First, although the simulator allows a significant number of tests, Algorithm~\ref{alg:qnt} terminates within a manageable testing effort throughout all cases in this section. For example, consider the three OSSs with Digit-HZDRL involved ($O_1$, $O_2$, and $O_3$) with executing Algorithm~\ref{alg:qnt} for 15 times (5 random seeds for each OSS). The average number of runs of scenarios is 619($\pm 52$) including on average 116($\pm32$) runs terminated in $\mathcal{C}$ or failed at initialization. Each run of a scenario lasts for at most 10 seconds. Note that given $\epsilon=0.02$ and $\beta=0.001$, the validation process takes $342$ samples which occupies over half of the total testing effort. The testing effort can be further reduced by relaxing the required confidence level, the probability, and the resolution coefficients. Another factor that affects the computational complexity is the Curse-of-Dimensionality. The cardinality of $\Phi_c$ grows exponentially with the OSS dimension. In practice, we have configured $\delta$ to be an appropriate value as the trade-off between the resolution precision and the admissible testing effort. 

Second, throughout all experiments in this section, the initial candidate set is mostly a significant over-approximation of the obtained almost safe set. In practice, with more expert experience and domain-knowledge involved, the initialization set can be chosen wisely, and the sampling efficiency can be naturally improved as well.

Finally, the role of an observable state and other uncertainties are interchangeable as part of the OSS design. For example, the actual acceleration state within each mode, the surface friction, and the contact phase are considered as part of the disturbances and uncertainties $W$ throughout all experiments and are assumed to follow a certain unknown but fixed distribution determined by the testing execution and environment. However, in some of the testing configurations, especially with simulators, the aforementioned states can be controllable and observable, hence can also serve as added dimensions to the OSS, yet the computational complexity would also increase per the first remark.

\section{Conclusion}\label{sec:conclusion}
This paper has presented, to the best of our knowledge, the first data-driven, scenario-based safety testing framework in the legged robot regime. In particular, we have presented a theoretically sound and empirically effective scenario-sampling framework with few to none assumptions to test, validate, and characterize the safety performance of various legged robot locomotion controllers. It is of future interest to explore more efficient quantification algorithms and to establish community-wide accepted OSS designs for legged robots.

\bibliography{output}
\bibliographystyle{IEEEtran}

\end{document}